\documentclass[10pt,twocolumn,letterpaper]{article}

\usepackage[pagenumbers]{cvpr}      %

\usepackage{cuted} %

\usepackage{currfile} %

\usepackage{caption} %

\DeclareMathOperator*{\argmax}{arg\,max}
\DeclareMathOperator*{\argmin}{arg\,min}

\usepackage{multirow}

\usepackage{graphicx}
\usepackage{booktabs}

\definecolor{cvprblue}{rgb}{0.21,0.49,0.74}
\usepackage[pagebackref,breaklinks,colorlinks,allcolors=cvprblue]{hyperref}

\def\confName{CVPR}
\def\confYear{2027}

\newcommand{\method}{EgoJev\xspace}
\usepackage{xspace}

\title{\method: Finding Evidence in Egocentric Video\\by Reading VLM Decisions}

\author{Chunggi Lee \qquad Hanspeter Pfister\\
Harvard University}

\usepackage{placeins}
\begin{document}
\twocolumn[{%
\renewcommand\twocolumn[1][]{#1}%
\maketitle
\begin{center}
  \includegraphics[width=\linewidth]{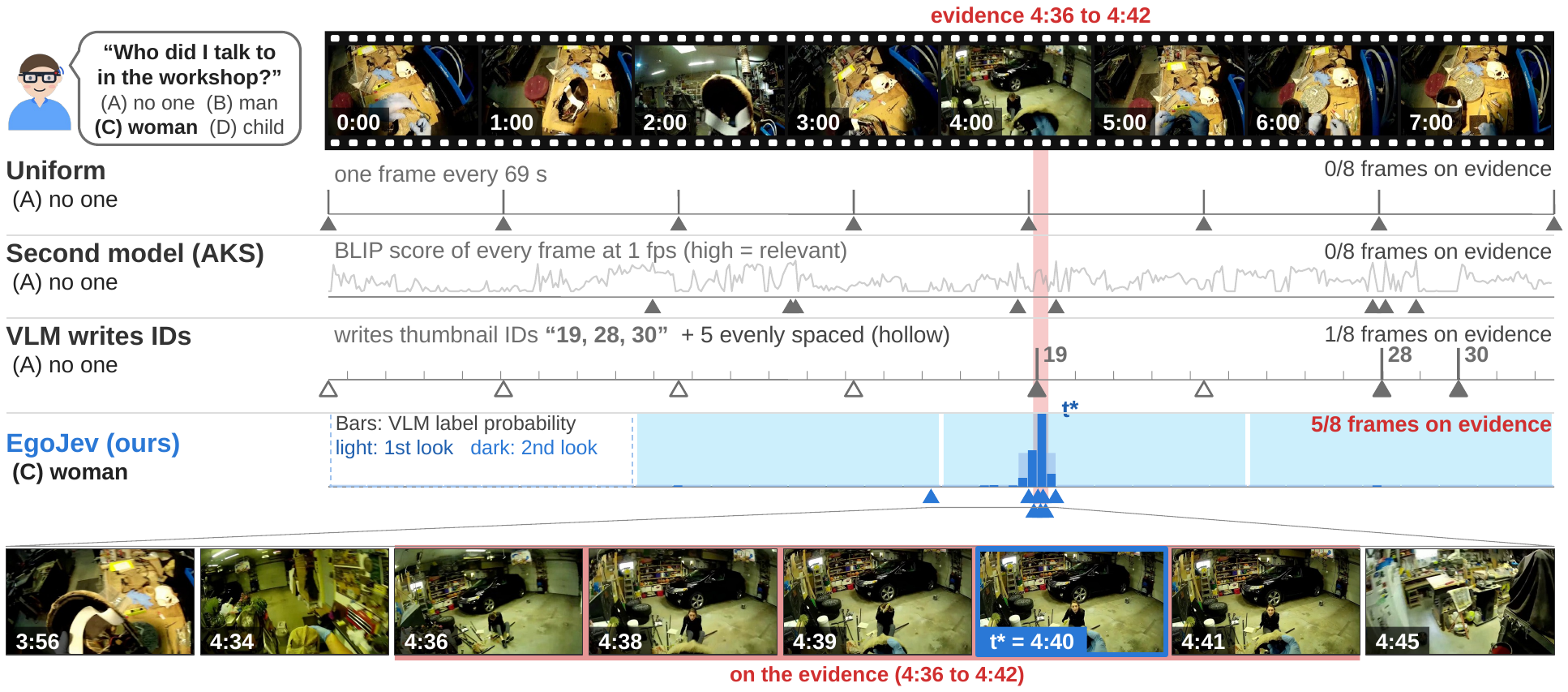}
  \captionof{figure}{\textbf{Four ways to choose 8 frames} for one question on an 8-minute video (Qwen3.5-4B, QAEgo4D). Uniform sampling, a second model (AKS) and written IDs put at most one of their frames (triangles) inside the brief evidence window (red band), and all three answer wrongly. \method reads the VLM's choice over labeled thumbnails as a distribution (bars, square-root scale), looks again in the windows it points to (shaded; dashed: skipped), and places its frames at the quantiles. Five of its 8 frames lie inside the evidence window. It answers correctly and returns the evidence time $t^*$ without generating text.}
  \label{fig:teaser}
\end{center}
}]

\begin{abstract}
Questions about long egocentric video often depend on one brief moment, such as the second a hand puts an object down.
A vision-language model (VLM) that sees a few evenly spaced frames usually misses it.
Training-free methods rank frames with a separate image-text model or ask the VLM to write a time.
In a controlled test, the VLM itself locates the moment better when it chooses among labeled frames, and we read this choice from its output distribution.
With the same candidates, this choice lands on the evidence more often than image-text scores or a written time span, and it compares all candidates in one decision.
Based on this finding, we propose \method, a training-free frame selector for wearable assistants.
\method reads where the evidence is from labeled thumbnails and looks again at the windows it points to.
It places its few answer frames at the quantiles of the resulting distribution.
One frozen VLM selects, answers and returns the evidence time, with no second model.
A wearable device can encode its coarse thumbnails while it records.
With Qwen3.5-4B and 8 answer frames, \method raises QAEgo4D accuracy from 51.2\% with uniform sampling to 62.0\%, and it outperforms recent training-free frame selectors on QAEgo4D and MyEgo.
A single-level version runs on a Mac and an iPhone.
\end{abstract}

\section{Introduction}
\label{sec:intro}

Smart glasses can record what their wearer sees and later answer questions about it, such as \textit{``Where did I put the scissors?''} or \textit{``What did I take out of the fridge?''}~\cite{ego4d,qaego4d}.
Such an assistant works under three limits.
First, it can afford only a few frames per question. The device has little compute, and sending video to a server costs bandwidth and privacy.
Second, a single small model must do the work, since a second network that watches every frame adds memory and power.
Third, the answer should come with the moment it was taken from, because a user who asks about the past wants to see that moment.

Training-free methods choose better frames in three ways, and none fits these limits well.
Similarity-based selectors~\cite{aks,bolt} score every frame with a separate image-text model such as CLIP~\cite{clip} or BLIP~\cite{blip}.
They need that second model, and they only compare frames with the question text.
A second way asks the VLM to rate each frame on its own, for example by the probability of ``Yes''~\cite{sevila}.
This needs one forward pass per frame, and frames scored in separate calls are never compared.
The third way asks the VLM to write when the evidence appears, as a time span or as frame IDs~\cite{tcot}.
The spans it writes land on the evidence less often.

Our idea is to let the VLM choose and to read its choice from the output probabilities (\cref{fig:teaser}).
The VLM sees the candidate frames as separate images, each after a number label such as ``Segment~3:'', and is asked where the answer is.
One forward pass then gives the probability of each label.
This treats the VLM as a decision model that returns a probability for each option, as in Jev~\cite{jev}.
Based on this idea, we propose \method, a training-free frame selector for wearable assistants.
\method finds the evidence in two steps.
It first reads one small labeled thumbnail per video segment.
Brief evidence often falls between two thumbnails, so it then reads again inside the windows with the most probability and combines the two levels.
It then places the answer frames at the quantiles of this distribution, so they gather when the model is sure and spread when it is not.
The same VLM answers from these frames, and one more readout returns the evidence time.
Unlike per-frame rating~\cite{sevila} or similarity selectors~\cite{aks,bolt}, \method compares all candidates in one decision. Unlike ID writing~\cite{tcot}, it reads this decision as a distribution and generates no text.
One frozen VLM thus selects, answers and points to its evidence, with no second model.
\method is also plug-and-play, since it needs only the output probabilities of the VLM that answers.
Because the thumbnails come before the question, a wearable device can encode them while it records.

We evaluate \method on three egocentric benchmarks with four VLMs from three families.
On the same candidates, reading the choice finds the evidence more often than rating frames, writing spans or scoring with BLIP and CLIP.
It also works much better with labeled images than with one video.
Its lead is largest on evidence shorter than 3\,s, the kind of moment that egocentric questions ask about.

In summary, our contributions are:
\begin{itemize}
  \item We find that a frozen VLM (Qwen3.5-4B) localizes brief evidence in egocentric video better when it chooses among labeled frames and we read its decision from the output distribution. With the same model and candidates, this beats rating each frame and writing a span in less than a third of their localization time.
  \item We propose \method, a training-free and plug-and-play frame selector for wearable assistants. It reads the VLM's decision from coarse to fine and places the answer frames at its quantiles. One frozen VLM selects, answers and returns the evidence time, with no second model. Its thumbnails can be encoded during recording.
  \item With Qwen3.5-4B, \method raises QAEgo4D accuracy from 51.2\% to 62.0\% and outperforms recent training-free frame selectors on QAEgo4D and MyEgo. Its mean accuracy exceeds that of AKS, BOLT, FOCUS and Write IDs with each of the four VLMs in \cref{tab:main,tab:models}, and a single-level version runs on a Mac and an iPhone.
\end{itemize}

\section{Related Work}
\label{sec:related}

\paragraph{Frame selection for video VLMs.}
A VLM can read only a few frames of a long video, so many training-free methods decide which frames to show it.
Most of them score frames by image-text similarity.
AKS~\cite{aks} balances relevance and coverage with BLIP~\cite{blip} scores.
BOLT~\cite{bolt} samples frames in proportion to sharpened CLIP scores, with a sharpness tuned per benchmark.
Others add multi-resolution inputs~\cite{qframe}, bandit search~\cite{focus}, answer options~\cite{vqos}, or scene changes~\cite{tasker}.
These methods run a second embedding model over every candidate frame.
Other methods read attention maps inside a second, smaller MLLM whose layer and budget are tuned on held-out videos~\cite{attnselect} or let an LLM pick frames from generated captions~\cite{videoagent,videotree,llovi}.
EcoFrame~\cite{ecoframe} reads the attention and output entropy of the answering VLM to decide when to add frames, and still scores relevance with CLIP.
A few methods let a VLM guide the search.
T*~\cite{tstar} searches grids of frames with an object detector, and A.I.R.~\cite{air} scores frames at 1\,fps with CLIP and then has a VLM write a score for each candidate, for up to six rounds.
\method targets egocentric questions whose evidence is brief. In contrast to these methods, it needs no second model and generates no text: the answering VLM sees all candidates in one prompt, and we read one distribution over them.

\paragraph{Temporal grounding.}
Egocentric benchmarks ask about brief past moments~\cite{ego4d,qaego4d} or about whole clips~\cite{egoschema}.
The first kind rewards finding the moment, and its evidence annotations let us measure localization directly.
MyEgo~\cite{myego} asks about the camera wearer and marks the answer moment.
Video LLMs can be trained to write timestamps~\cite{timechat,vtimellm}, and GroundVQA~\cite{groundvqa} trains a model to answer and ground egocentric questions together.
NumPro~\cite{numpro} draws frame numbers on the frames so that a model can write the number of the relevant frame, and TimeMarker~\cite{timemarker} interleaves time tokens with frames during training.
TCoT~\cite{tcot} shows the VLM segments of 64 frames with IDs and has it write the IDs of the relevant ones, one call per segment.
These methods output time or frame IDs as generated text, and most need training.
\method is training-free and reads the evidence time from label probabilities.

\paragraph{Decision readout.}
Reading a choice from label probabilities is an old way to use a language model as a classifier~\cite{calibrate}.
Jev~\cite{jev} recently framed it as a decision model that returns a probability for each option in one forward pass, and open frozen LLMs match it without training~\cite{innerjev,llm2jev}.
Label wording strongly shapes these decisions~\cite{labelsoverride}.
Listwise rerankers write an order of passage IDs~\cite{rankgpt}, and FIRST~\cite{first} reads the ranking from the logits of the first output token.
For video, SeViLA~\cite{sevila} scores each frame by the probability of ``Yes'' to whether it helps answer the question, and UG-Sample~\cite{ugsample} by the entropy of the answering MLLM's output given that frame.
Each frame is then scored in its own forward pass and is never compared with the others.
We read one decision over all candidate images and use its whole distribution to place the frame budget.
One VLM thus makes three decisions: where the evidence is, what the answer is, and when it was seen.

\section{Method}
\label{sec:method}

\begin{figure*}[t]
  \centering
  \includegraphics[width=\linewidth]{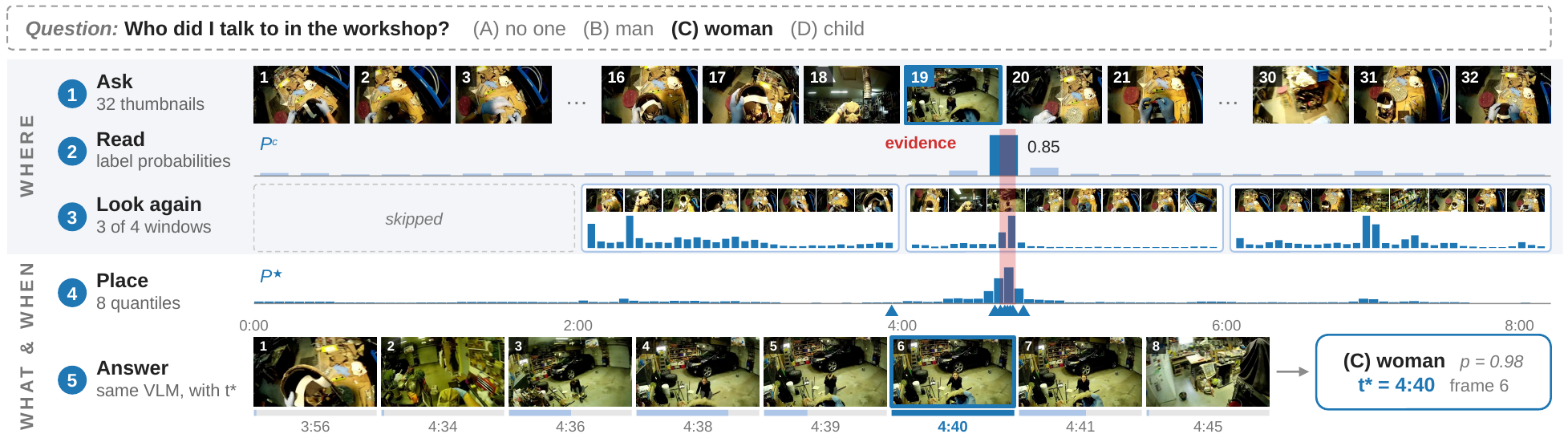}
   \caption{\textbf{Overview of \method} on one video (Qwen3.5-4B, QAEgo4D). (1) The VLM sees 32 labeled thumbnails with the question and options. (2) \method reads the probability of each label (bars on a square-root scale, largest 0.85, red band: annotated evidence). (3) The VLM reads 32 denser thumbnails in each of the 3 windows with the most probability. (4) The two readouts give $P^{\star}$, and the 8 answer frames are its quantiles (triangles). (5) The same VLM answers from these frames ($p$: probability of the chosen answer) and reads which frame shows the evidence (bars under the frames), which gives $t^*$. No step generates text.}
  \label{fig:method}
\end{figure*}

Our goal is to answer a multiple-choice question about a long egocentric video with a frozen vision-language model (VLM), a fixed frame budget, and no training.
The answer usually depends on a short moment, while uniform sampling spends most of the budget elsewhere.
\method asks one VLM for three decisions and reads each one from its output probabilities: \emph{where} the evidence is (\cref{sec:localize}), \emph{what} the answer is (\cref{sec:answer}), and \emph{when} the evidence was seen (\cref{sec:when}), as shown in \cref{fig:method}.
The key step is the first one: its probabilities, used as a whole distribution, decide where the frame budget goes.
One distribution drives every later step: it chooses which windows to look at again, decides how many answer frames go where, and these frames are read once more for the evidence time.

\subsection{Problem setup}
\label{sec:prelim}
\label{sec:setup}

A video $V$ comes with a question $q$ and options $O=\{o_1,\dots,o_K\}$.
The VLM can see only $B$ frames, far fewer than the video has, and by default they are sampled uniformly.
A vision encoder and a projector turn each frame into visual tokens, and the language model with parameters $\theta$ reads them with the text as one input $X$.
With a fixed budget $B$, the answer depends on which frames are selected.
Similarity-based methods~\cite{aks,bolt} score each frame by its cosine similarity to the question in a separate image-text model, and generation-based methods ask the VLM to write a time or a frame ID.
\method scores the frames with the output probabilities of the VLM that answers.

\subsection{Where: finding the evidence}
\label{sec:localize}
\label{sec:readout}

The central step of \method asks the VLM where the evidence is, as one multiple-choice question over candidate frames.
Given candidate frames $f_1,\dots,f_N$, each frame is shown as its own image after a number label, followed by the question, the options and an instruction:
\begin{equation}
  X_{\mathrm{loc}} = \big[\,\ell_1, v_1, \dots, \ell_N, v_N,\; q,\; O,\; u,\; \text{``Segment: [''}\,\big],
  \label{eq:xloc}
\end{equation}
where $\ell_i$ is the label ``Segment~$i$:'', $v_i$ are the visual tokens of $f_i$, and $u$ is \textit{``Which segment shows what is needed to answer this question?''}
The options help the model look for the object or action that tells them apart.

\paragraph{Reading the choice.}
The VLM would answer this prompt by writing a number, but before it writes, it already gives a probability to every possible next token.
Let $p_\theta(\cdot \mid X)$ be these probabilities after the input $X$.
The answer completes ``Segment: ['', so label $i$ is scored as the tokens of ``$i$]''. A label such as ``17]'' is several tokens (\eg, ``1'', ``7'', ``]''), so its probability is the product of the probabilities of its tokens $c_{i,1},\dots,c_{i,L_i}$, each given the tokens before it.
The model also gives some probability to tokens that are not labels, so these products are normalized over the $N$ labels:
\begin{equation}
  P(i) \;\propto\; \prod_{l=1}^{L_i} p_\theta\big(c_{i,l} \mid X_{\mathrm{loc}}, c_{i,<l}\big), \qquad \sum_{i=1}^{N} P(i) = 1.
  \label{eq:numeric}
\end{equation}
This readout $P(i)$ is the model's choice among the $N$ frames, as a distribution.
Each label ends with ``]'' so that the probability of ``1]'' does not include ``17]''.
This takes one forward pass plus a few cached steps per label, writes no text, and uses only these output probabilities, not hidden states or attention.

Because the model sees all candidates before it answers, it can compare them.
Methods that rate each frame on its own, such as SeViLA~\cite{sevila}, score the frames in separate calls, so the frames are never compared and their scores need not share a scale.
Here all candidates share one distribution.
The frames must enter as separate labeled images: a video encoder may merge neighboring frames, and then a single frame has no label to choose (\cref{sec:exp_format}).

\paragraph{Coarse to fine.}
\method first splits the video into $N$ equal segments and shows each by its middle frame as a small thumbnail of a few dozen visual tokens.
The readout over these thumbnails is the coarse distribution $P^{c}_j$, $j=1,\dots,N$.
Brief evidence is often shorter than a segment, so it can fall between two thumbnails.
\method therefore takes a second, closer look where $P^{c}$ points.
For this, the video is also split into $W$ windows, and each window is divided again into $N$ fine segments $s$ with their own thumbnails.
For the $k$ windows with the most coarse probability, \method reads these new, denser thumbnails with the same prompt.
This gives the fine distribution $P^{f}_s$ over all fine segments of the video.
In each of these $k$ windows, $P^{f}$ is the new readout, scaled so that the window keeps its coarse probability in total.
In the other $W-k$ windows, $P^{f}$ spreads the coarse probability evenly over its fine segments.
The two levels are combined by a geometric mean,
\begin{equation}
  P^{\star}_s \;\propto\; \sqrt{P^{c}_{j(s)}\, P^{f}_s},
  \label{eq:c2f}
\end{equation}
where $j(s)$ is the coarse segment that contains $s$, and $P^{\star}$ is normalized over all fine segments.
Both levels are readouts of the answering VLM, unlike FOCUS~\cite{focus}, which refines with BLIP scores, and T*~\cite{tstar}, which zooms in with a detector.
The square root averages the two levels in log space. A plain product would count the evidence twice and make $P^{\star}$ sharper (\cref{sec:exp_ablation}).
With $k=0$ this is the single-level readout.

\subsection{What: from the readout to the answer}
\label{sec:answer}

\begin{table*}[!t]
  \centering
  \small
  \caption{\textbf{Accuracy and time} (Qwen3.5-4B, 8 answer frames). Avg.: mean of the three datasets. \# frames: frames the selector reads per question, with $T$ the video length in seconds. Time: seconds per question on QAEgo4D. On EgoSchema \method uses the label check (\cref{sec:label}). In all tables, best is bold and second best is underlined.}
  \label{tab:main}
  \resizebox{0.85\linewidth}{!}{%
  \begin{tabular}{@{}lccccccc@{}}
    \toprule
    Method & \# frames & Extra model & Time (s) & QAEgo4D & MyEgo & EgoSchema & Avg. \\
    \midrule
    Uniform & -- & none & 1.7 & 51.2 & 36.1 & \textbf{65.2} & 50.8 \\
    AKS~\cite{aks} & 1\,fps${\times}T$ & BLIP & 90.7 & 56.4 & 39.4 & 60.4 & 52.1 \\
    BOLT~\cite{bolt} & 1\,fps${\times}T$ & CLIP & 52.9 & 54.2 & 41.3 & 63.6 & 53.0 \\
    A.I.R.~\cite{air} & 1\,fps${\times}T$ & CLIP & 82.9 & \underline{60.8} & \underline{41.9} & 63.4 & \underline{55.4} \\
    Write IDs~\cite{tcot} & 32 & none & 7.5 & 55.8 & 39.8 & \underline{64.8} & 53.5 \\
    FOCUS~\cite{focus} & 230 & BLIP & 49.0 & 56.2 & 40.3 & 62.2 & 52.9 \\
    \midrule
    \method & 128 & none & 26.7 & \textbf{62.0} & \textbf{42.6} & \textbf{65.2} & \textbf{56.6} \\
    \bottomrule
  \end{tabular}}
\end{table*}

$P^{\star}$ says how strongly the VLM points to each segment, so \method places the $B$ answer frames where it points.
Spreading each $P^{\star}_s$ evenly over its segment gives a distribution $\mu$ over time, with cumulative distribution $F$.
\method places the $b$-th answer frame at
\begin{equation}
  t_b = F^{-1}\!\Big(\frac{b - 1/2}{B}\Big), \qquad b=1,\dots,B.
  \label{eq:qp}
\end{equation}
Each frame covers an equal share $1/B$ of the probability and sits in the middle of its share.
Among frames that each stand for $1/B$ of the probability, these minimize the average time between an evidence moment drawn from $\mu$ and the frame that stands for it, which is the 1-Wasserstein distance~\cite{xu2019best} (proof in \cref{sec:app_proof}):
\begin{equation}
  \{t_b\} = \argmin_{t_1,\dots,t_B} \mathcal{W}_1\Big(\mu,\; \frac{1}{B}\sum_{b=1}^{B} \delta_{t_b}\Big),
  \label{eq:quant}
\end{equation}
where $\delta_{t_b}$ is the point mass of a frame at time $t_b$.

The placement rule adds no threshold and no per-dataset sharpness.
When the model is sure, most frames land in one segment and cover the brief moment densely.
When it is unsure, they spread over the likely segments.
When $P^{\star}$ is flat, the frames are evenly spaced, so even spacing is the special case with no information.
BOLT~\cite{bolt} also samples by inverse transform, but on CLIP similarities that are made into a distribution with a sharpness $\alpha$ set per dataset, while $P^{\star}$ is already normalized over the candidates.
A temperature $\tau$ in $(P^{\star})^{1/\tau}$ could make the frames more or less concentrated, and $\tau=1$ is used for every model and dataset.
The answer frames use the answer-frame cap of \cref{sec:exp_setup}, each after the label ``Frame~$b$:''.
The answer is read the same way, with the option letters as labels.
Each option $y$ is shown with a letter $c_y$ (\eg, ``A''), and the answer is the most likely letter,
\begin{equation}
  \hat{y} = \argmax_{y} \; p_\theta(c_y \mid X_{\mathrm{ans}}),
  \label{eq:letter}
\end{equation}
where $X_{\mathrm{ans}}$ holds the labeled answer frames, the question and the options.

\subsection{When: the evidence time}
\label{sec:when}

An assistant should also show the moment it used.
After the answer, \method adds \textit{``Which frame shows what is needed to answer the question?''} to the same context and reads the probabilities of the $B$ frame labels as in \cref{eq:numeric}.
The evidence time $t^*$ is the time of the most likely frame.
This adds 41 text tokens and no visual tokens.

\paragraph{Cost.}
\method runs $1+k$ localization passes and one answer pass per question, and generates no text.
Because the thumbnails come before the question, the coarse thumbnails can be encoded and cached while the video is recorded, so coarse localization at question time only reads the question.

\section{Experiments}

\label{sec:experiments}

\cref{sec:exp_main,sec:exp_models} compare \method with uniform sampling and recent training-free frame selectors on four VLMs.
\cref{sec:exp_ask,sec:exp_ablation} then test the idea behind it and each of its parts, in the order of \cref{sec:method}.
Last, \cref{sec:exp_ground,sec:exp_device} measure the evidence time and the cost on a device.

\begin{table}[!t]
  \centering
  \small
  \caption{\textbf{Same candidates, and \method without options} (Qwen3.5-4B, 8 answer frames). AKS, BOLT and Write IDs choose from the 32 thumbnails that \method reads first. The last two rows read all 128 thumbnails. \method without options is read without the label check.}
  \label{tab:same}
  \resizebox{\linewidth}{!}{%
  \begin{tabular}{@{}lccc@{}}
    \toprule
    Method & QAEgo4D & MyEgo & EgoSchema \\
    \midrule
    AKS, same 32 thumbnails & 54.8 & 40.0 & 63.2 \\
    BOLT, same 32 thumbnails & 53.0 & 39.3 & 62.6 \\
    Write IDs, same 32 thumbnails & 55.8 & 39.8 & 64.8 \\
    \midrule
    \method single level, same 32 thumbnails & 57.2 & 42.2 & \underline{65.2} \\
    \method without options & \underline{59.0} & \underline{42.5} & \textbf{65.8} \\
    \method & \textbf{62.0} & \textbf{42.6} & \underline{65.2} \\
    \bottomrule
  \end{tabular}}
\end{table}

\begin{table*}[!t]
  \centering
  \small
  \caption{\textbf{Other VLMs} (8 answer frames, same settings as \cref{tab:main}). QA: QAEgo4D, My: MyEgo, ES: EgoSchema, Avg.: mean of the three.}
  \label{tab:models}
  \resizebox{\linewidth}{!}{%
  \begin{tabular}{@{}l*{3}{cccc@{\hspace{8pt}}}@{}}
    \toprule
     & \multicolumn{4}{c}{Qwen3-VL-8B~\cite{qwen3vl}} & \multicolumn{4}{c}{MiniCPM-V-4.5~\cite{minicpmv45}} & \multicolumn{4}{c}{GLM-4.1V-9B~\cite{glm41v}} \\
    \cmidrule(r{8pt}){2-5}\cmidrule(r{8pt}){6-9}\cmidrule(r{8pt}){10-13}
    Method & QA & My & ES & Avg. & QA & My & ES & Avg. & QA & My & ES & Avg. \\
    \midrule
    Uniform & 55.4 & 36.1 & \underline{68.8} & 53.4 & 53.6 & 34.9 & \underline{53.8} & 47.4 & 53.0 & 33.1 & 61.8 & 49.3 \\
    AKS (1\,fps) & \underline{60.0} & 40.7 & 66.0 & 55.6 & \underline{59.6} & \underline{39.8} & 49.2 & 49.5 & \underline{57.4} & \textbf{39.1} & 59.0 & 51.8 \\
    BOLT ($\alpha=2.5$) & 54.0 & 40.4 & \underline{68.8} & 54.4 & 57.4 & 39.3 & \textbf{54.0} & 50.2 & 53.2 & 37.7 & \textbf{63.0} & 51.3 \\
    Write IDs & 57.8 & \underline{42.8} & \underline{68.8} & \underline{56.5} & 56.8 & 38.6 & 51.6 & 49.0 & 55.8 & 36.9 & 61.6 & 51.4 \\
    FOCUS & 58.2 & 40.8 & 66.0 & 55.0 & \textbf{61.2} & 38.7 & 51.4 & \underline{50.4} & 56.4 & \underline{39.0} & 61.6 & \underline{52.3} \\
    \midrule
    \method & \textbf{60.4} & \textbf{44.0} & \textbf{69.0} & \textbf{57.8} & 57.2 & \textbf{41.9} & \underline{53.8} & \textbf{51.0} & \textbf{58.0} & 38.9 & \underline{62.2} & \textbf{53.0} \\
    \bottomrule
  \end{tabular}}
\end{table*}

\subsection{Setup}
\label{sec:exp_setup}

\paragraph{Benchmarks.}
All three benchmarks are egocentric and multiple choice, which matches the wearable setting.
QAEgo4D~\cite{qaego4d} asks about objects and actions in Ego4D~\cite{ego4d} videos of about 8 minutes and marks the time window of the evidence.
Its multiple-choice test set, as released by GroundVQA~\cite{groundvqa}, has 500 questions over 148 videos.
MyEgo~\cite{myego} asks about moments long before the question and marks the answer moment.
The evaluation uses all 996 of its questions asked at least 15\,s after that moment, and the model sees the video only up to the question.
EgoSchema~\cite{egoschema} asks about whole 3-minute clips and has no evidence annotation.
The evaluation uses its public subset of 500 questions.

\paragraph{Baselines.}
\emph{Uniform} samples $B$ frames evenly, including the first and last frame.
\emph{AKS}~\cite{aks} scores every frame at 1\,fps with BLIP and balances relevance and coverage.
\emph{BOLT}~\cite{bolt} scores every frame at 1\,fps with CLIP and samples frames in proportion to the sharpened scores.
\emph{A.I.R.}~\cite{air} filters frames with CLIP and lets the VLM rate them over up to six rounds.
\emph{FOCUS}~\cite{focus} searches clips as bandit arms with BLIP scores.
\emph{Write IDs} lets the same VLM write the labels of the relevant thumbnails among the 32 that \method reads first, as TCoT~\cite{tcot} does, with up to 256 generated tokens.
Each written label places a frame in its segment, and evenly spaced frames fill the rest of the 8.
AKS, BOLT, A.I.R.\ and FOCUS run with their official code, settings and queries.

\paragraph{Metrics.}
Accuracy is the share of correct answers. For localization, a hit means that the top segment overlaps the evidence, and a time counts as correct within 8\,s of the evidence.

\paragraph{Implementation details.}
The main VLM is Qwen3.5-4B~\cite{qwen35}, kept frozen, and the same VLM answers for every method.
Every method answers from $B=8$ labeled frames with the same cap on visual tokens per frame: 384 on QAEgo4D, 88 on EgoSchema and 72 on MyEgo.
Apart from these shared caps, every method keeps its own settings fixed across datasets and models.
Time is the mean end-to-end time per question over 50 QAEgo4D questions on one A100 MIG slice, including frame scoring.
\method uses $N=32$ thumbnails, $W=4$ windows, $k=3$ refined windows and $\tau=1$.
\label{sec:label}\label{sec:exp_label}
VLMs can favor some option labels whatever the content~\cite{zheng2024mcq}.
When a short test on 40 questions shows that the coarse readout follows the label numbers, it is also read in reverse order and the two are combined by their geometric mean, as in permutation debiasing~\cite{zheng2024mcq}. We call this the label check.
With all four VLMs this fires only on EgoSchema.
AKS uses its official candidates and default thresholds, and BOLT, which has no setting for QAEgo4D or MyEgo, uses $\alpha=2.5$, its setting for EgoSchema.

\begin{figure}[t]
  \centering
  \includegraphics[width=0.72\linewidth]{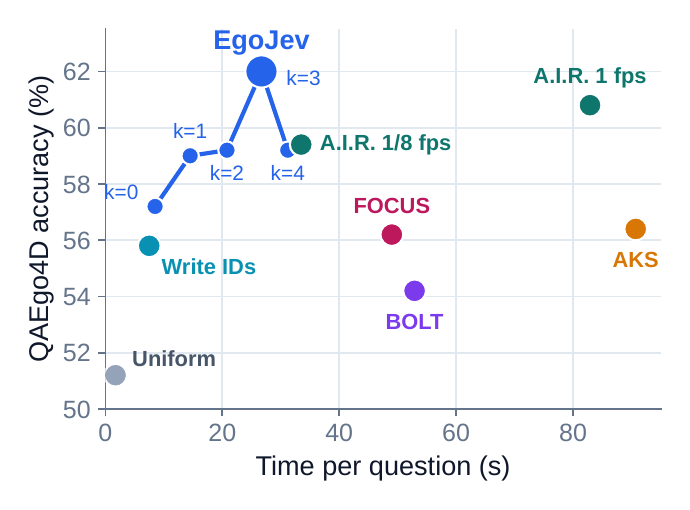}
  \caption{\textbf{Accuracy vs.\ time per question} (QAEgo4D, Qwen3.5-4B, 8 answer frames, one A100 MIG slice). Baselines at the settings of \cref{tab:main}, and A.I.R.\ also at 1/8\,fps (59.4\% in 33.5\,s). The EgoJev line refines $k=0$ to 4 windows. Time: mean end-to-end time over 50 questions.}
  \label{fig:acctime}
\end{figure}

\subsection{Main results}
\label{sec:exp_main}

\cref{tab:main} compares choosing with the answering VLM against choosing with a separate scorer.
\method is the most accurate on QAEgo4D and MyEgo (62.0\% and 42.6\%).
It is also 3.1 times faster per question than A.I.R., which uses a second model.
On EgoSchema, which asks about whole clips, every selector that uses a second model falls below uniform sampling, while \method with the label check matches it.
Given the same 32 thumbnails, AKS, BOLT and Write IDs all stay below single-level \method on all three datasets (\cref{tab:same}).
Reading the choice thus beats both similarity scores and written IDs.
Without options in its prompt, \method still leads AKS and BOLT (\cref{tab:same}), and options do not help AKS or BOLT on QAEgo4D (56.0\% and 52.6\%, \cref{sec:app_same}).
\method also puts an answer frame within 8\,s of the evidence for 67.2\% of QAEgo4D and 66.7\% of MyEgo questions, against at most 59.2\% and 58.0\% for the baselines. Its lead over the best baseline is largest when the evidence lasts less than 3\,s (62.6\% against 58.0\%, \cref{sec:app_more}).

\paragraph{Accuracy and time.}
\cref{fig:acctime} plots accuracy against the end-to-end time per question.
\method reaches 62.0\% in 26.7\,s, while AKS and BOLT take 53 to 91\,s at their published settings.

\subsection{Other VLMs}
\label{sec:exp_models}

To test whether the gains hold beyond one model, \cref{tab:models} runs uniform sampling, AKS, BOLT, Write IDs, FOCUS and \method on VLMs from three families with the settings of \cref{tab:main}.
Over the three datasets, the mean accuracy of \method is higher than the better of AKS and BOLT by 2.2 points with Qwen3-VL-8B, 1.2 with GLM-4.1V-9B and 0.8 with MiniCPM-V-4.5 (3.6 with Qwen3.5-4B).
It is also higher than that of FOCUS and Write IDs with all three.
With Qwen3-VL-8B it is above AKS and BOLT on every dataset.

\begin{table}[!t]
  \centering
  \small
  \caption{\textbf{How to ask.} Segment hit rate with the same 16 candidate frames (one per segment). The first three rows use the same VLM (Qwen3.5-4B). Time: localization seconds per question on QAEgo4D.}
  \label{tab:ask}
  \resizebox{\linewidth}{!}{%
  \begin{tabular}{@{}lcccc@{}}
    \toprule
    Scorer over 16 candidates & QAEgo4D & MyEgo & Passes & Time (s) \\
    \midrule
    VLM, choose one (\method) & \textbf{28.0} & \textbf{18.5} & 1 & \underline{0.44} \\
    VLM, rate each (``Yes'') & 18.2 & \underline{13.5} & 16 & 1.41 \\
    VLM, write a span & 15.8 & 12.6 & 1 + decode & 1.68 \\
    \midrule
    BLIP image-text matching (AKS) & \underline{19.0} & 12.9 & 16 & 1.32 \\
    CLIP similarity (BOLT) & 15.6 & 12.0 & 16 & \textbf{0.13} \\
    Random segment & 8.9 & 6.2 & -- & -- \\
    \bottomrule
  \end{tabular}}
\end{table}

\begin{table}[!t]
  \centering
  \small
  \caption{\textbf{Coarse-to-fine depth} $k$ (Qwen3.5-4B, number of refined windows out of 4, 8 answer frames, without the label check). Time: end-to-end per question, as in \cref{tab:main}. +32 at quantiles: a second readout over 32 thumbnails placed at the quantiles of $P^{c}$ instead of in windows.}
  \label{tab:depth}
    \setlength{\tabcolsep}{2.2pt}
  \footnotesize
  \begin{tabular}{@{}lccccc@{}}
    \toprule
    $k$ & QAEgo4D & MyEgo & EgoSchema & Mean & Time (s) \\
    \midrule
    0 (1 level) & 57.2 & 42.2 & \textbf{65.2} & 54.9 & 8.5 \\
    1 & 59.0 & 41.9 & 64.2 & 55.0 & 14.5 \\
    2 & \underline{59.2} & 43.0 & \underline{65.0} & \underline{55.7} & 20.8 \\
    3 & \textbf{62.0} & 42.6 & 62.8 & \textbf{55.8} & 26.7 \\
    4 & \underline{59.2} & \underline{43.4} & 63.4 & 55.3 & 31.2 \\
    \midrule
    +32 at quantiles of $P^{c}$ & \underline{59.2} & 41.8 & 64.4 & 55.1 & 13.6 \\
    One readout, 128 & 58.6 & \textbf{43.5} & \underline{65.0} & \underline{55.7} & 26.4 \\
    3, plain product & 58.0 & \textbf{43.5} & 62.6 & 54.7 & 26.7 \\
    \bottomrule
  \end{tabular}
\end{table}

\subsection{How to ask the VLM where to look}
\label{sec:exp_ask}

\cref{tab:ask} asks whether the way the VLM is asked changes how well it finds the evidence.
Only the question changes: the VLM chooses one of the candidate frames, rates each frame alone by the probability of ``Yes'', or writes a time span.
A method finds the evidence when the segment it ranks first overlaps the evidence window.
The test uses 16 candidates, the middle frames of 16 equal segments.
Choosing finds the evidence segment in 28.0\% of QAEgo4D and 18.5\% of MyEgo questions.
Rating each frame alone finds it in 18.2\% and 13.5\%, and writing a span in 15.8\% and 12.6\%.
BLIP and CLIP, the scorers of AKS and BOLT, reach only 19.0\% and 15.6\% on QAEgo4D.
With the 32 candidates of \method, the order is the same on QAEgo4D: 28.6\% for choosing, against 21.0\% for rating, 17.8\% for a span, and 19.4\% and 18.8\% for BLIP and CLIP.

\paragraph{Input format.}
\label{sec:exp_format}
The readout needs the model to tell candidate frames apart.
In a simple test, one of 16 gray frames has a red square.
Given as labeled images, Qwen3.5-4B and Qwen2.5-VL-7B~\cite{qwen25vl} name the marked frame in 16 of 16 cases. Given as one video, they name it in only 3 and 2 cases.
On QAEgo4D, giving the 32 thumbnails as one video lowers single-level accuracy from 57.2\% to 53.2\% (\cref{sec:app_more}).

\begin{table}[!t]
  \centering
  \small
  \caption{\textbf{Frame layout} from the single-level readout (Qwen3.5-4B, 8 answer frames). Focus: 6 frames in the top segment and 2 uniform. Spread: the relevance and coverage rule of AKS on $P^{c}$. Strata: one frame per eighth of the video, placed by a second readout within that eighth.}
  \label{tab:layout}
  \resizebox{\linewidth}{!}{%
  \begin{tabular}{@{}lcccc@{}}
    \toprule
    Layout & QAEgo4D & MyEgo & EgoSchema & Mean \\
    \midrule
    Focus & \textbf{58.6} & 38.7 & 61.8 & 53.0 \\
    Spread & 56.2 & \textbf{42.4} & \underline{63.2} & 53.9 \\
    Strata & 56.6 & 40.4 & \textbf{65.2} & \underline{54.1} \\
    Quantiles (\method) & \underline{57.2} & \underline{42.2} & \textbf{65.2} & \textbf{54.9} \\
    \bottomrule
  \end{tabular}}
\end{table}

\begin{figure*}[t]
  \centering
  \includegraphics[width=\linewidth]{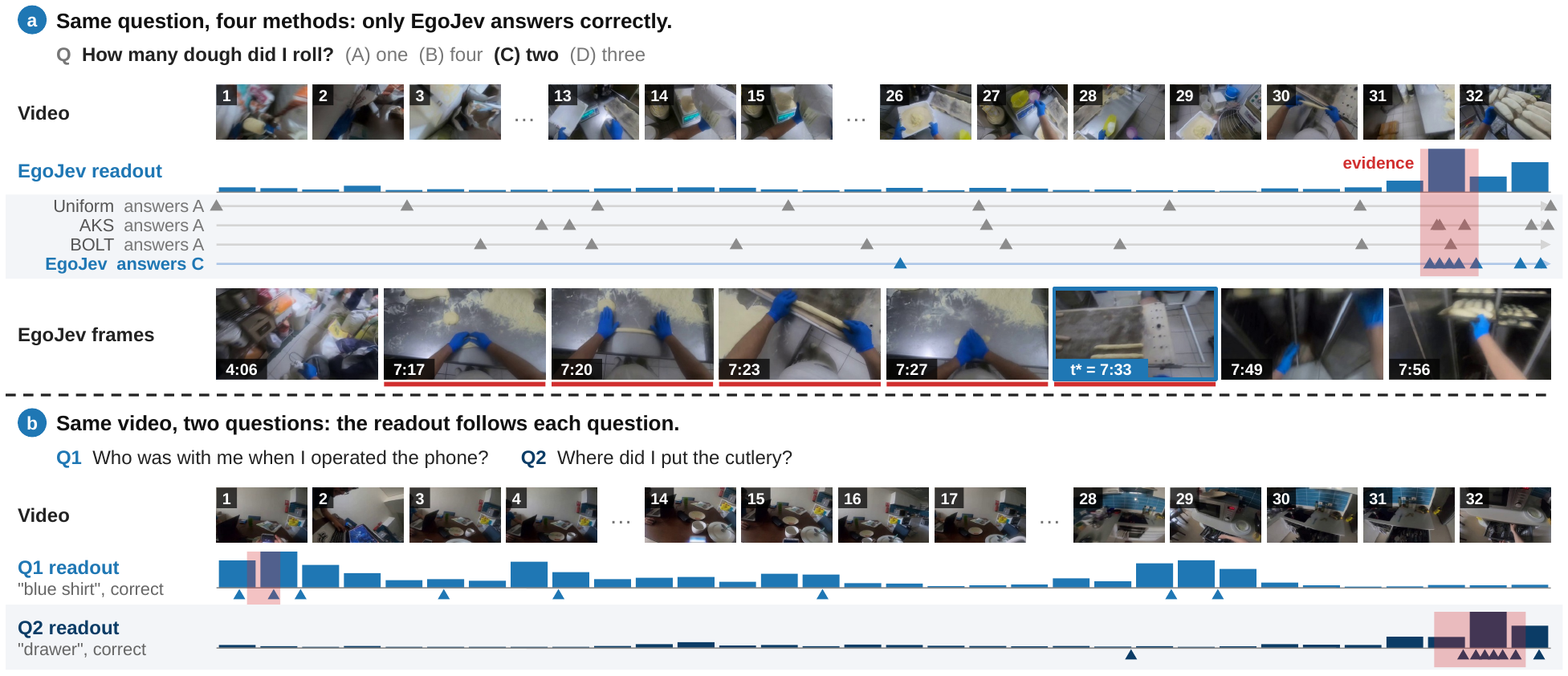}
  \caption{\textbf{Examples} (Qwen3.5-4B, QAEgo4D). (a) Four methods choose 8 answer frames (triangles) for one question. Only \method answers correctly, and 5 of its 8 frames lie inside the evidence window (red band; underlined frames). Its coarse readout $P^{c}$ over 32 segments (bars, square-root scale) peaks at this window. (b) For two questions about the same video, the readout (Q1 in blue, Q2 in dark blue) peaks at each question's evidence.}
  \label{fig:qual}
\end{figure*}

\subsection{Ablations}
\label{sec:exp_ablation}

\cref{tab:depth} varies how many windows are read at the fine level.
On QAEgo4D, accuracy rises by 4.8 points from a single level to three refined windows, the setting used everywhere.
A fourth window adds time and lowers the mean. One readout over 128 thumbnails is 3.4 points lower on QAEgo4D at the same mean and time.
Combining the two levels by a plain product instead of the geometric mean lowers QAEgo4D by 4.0 points and the mean by 1.1 (\cref{tab:depth}).
For the 54\% of QAEgo4D questions whose evidence no thumbnail shows, refinement raises the share with an answer frame within 8\,s of the evidence from 45.7\% to 54.6\% (\cref{sec:app_between}).

\paragraph{Frame layout.}
Quantile placement is a known rule, so we test whether it suits the readout.
\cref{tab:layout} starts from the single-level readout and changes where the 8 answer frames go.
Putting most frames in the top segment (Focus) helps QAEgo4D, where the evidence is brief, but costs 3.5 points on MyEgo and 3.4 on EgoSchema.
Spreading them out helps elsewhere: Spread is best on MyEgo and Strata ties on EgoSchema, but both lose on QAEgo4D.
The quantiles follow the readout: they gather when it is sure and spread when it is not, so they are first or second on every dataset and give the best mean.

\begin{table}[!t]
  \centering
  \small
  \caption{\textbf{Evidence time.} Share of questions whose single predicted time is within 8\,s of the evidence or inside it (QAEgo4D), and within 8\,s / 30\,s of the answer moment (MyEgo). Localization only: the time with the most probability in $P^{\star}$.}
  \label{tab:ground}
  \resizebox{\linewidth}{!}{%
  \begin{tabular}{@{}lcccc@{}}
    \toprule
     & \multicolumn{2}{c}{QAEgo4D} & \multicolumn{2}{c}{MyEgo} \\
    \cmidrule(lr){2-3}\cmidrule(l){4-5}
    Method & $\pm$8\,s & inside & 8\,s & 30\,s \\
    \midrule
    AKS (1\,fps, BLIP) & 27.8 & 18.8 & 21.3 & \underline{46.7} \\
    BOLT (1\,fps, CLIP) & 22.8 & 13.8 & 20.5 & 43.8 \\
    VLM, rate each & 14.2 & 8.4 & 17.6 & 39.3 \\
    VLM, write a span & 22.8 & 11.0 & 17.3 & 41.8 \\
    Random time & 5.9 & 2.7 & 9.9 & 30.9 \\
    \midrule
    \method, localization only & \underline{32.2} & \underline{21.6} & \underline{25.6} & 46.1 \\
    \method, evidence readout & \textbf{33.6} & \textbf{22.6} & \textbf{26.6} & \textbf{50.4} \\
    \bottomrule
  \end{tabular}}
\end{table}

\subsection{Evidence time}
\label{sec:exp_ground}

Besides the answer, an assistant should show the moment it used (\cref{sec:when}).
\cref{tab:ground} checks one time per method against the evidence, such as the frame it scores highest.
The evidence readout of \method costs 41 text tokens and lies within 8\,s of the evidence for 33.6\% of QAEgo4D questions, against 27.8\% for the frame that AKS scores highest.
In \cref{fig:qual}, only \method answers the first question correctly, and for two questions about one video its readout peaks at each question's evidence.

\subsection{On device}
\label{sec:exp_device}

A wearable assistant records continuously and answers on the device.
\begin{table}[!t]
  \centering
  \small
  \caption{\textbf{On device} (Mac, 4-bit weights, 6\,GB memory cap, medians over 20 questions). \method-S: single level. Rec.: compute while 8 minutes are recorded. Ans.: from question to answer. QA: QAEgo4D accuracy from GPU runs with the same frames. $^\ddagger$iPhone 15 Pro, answer frames capped at 192 tokens, 10 questions.}
  \label{tab:device}
  \setlength{\tabcolsep}{2pt}
  \footnotesize
  \begin{tabular}{@{}llcccc@{}}
    \toprule
    Method & Extra model & Rec.\ (s) & Ans.\ (s) & Total (s) & QA \\
    \midrule
    AKS, $B=8$ & BLIP 446M & 53.8 & 13.6 & 67.4 & 56.4 \\
    BOLT, $B=8$ & CLIP 428M & \underline{28.3} & \underline{10.4} & 38.7 & 54.2 \\
    \midrule
    \method-S, $B=8$ & none & \textbf{8.2} & 10.7 & \underline{18.8} & \underline{57.2} \\
    \method-S, $B=4$ & none & \textbf{8.2} & \textbf{5.9} & \textbf{14.0} & \textbf{59.6} \\
    \midrule
    \method-S, $B=4$, iPhone$^\ddagger$ & none & 18.3 & 8.0 & 26.2 & 58.0 \\
    \bottomrule
  \end{tabular}
\end{table}

To measure the cost there, every method runs on a Mac with 4-bit weights and a 6\,GB memory cap, in one process from recording to answer (\cref{tab:device}).
While the video is recorded, \method adds one thumbnail every 15\,s to the VLM's own cache, which costs 8.2\,s per 8 minutes.
At their published settings, BOLT and AKS run a second model on every frame at 1\,fps, which costs 28.3\,s and 53.8\,s.
A.I.R.\ and FOCUS also run a CLIP or BLIP scorer over many frames, and on the GPU they take 82.9\,s and 49.0\,s per question, against 26.7\,s for \method (\cref{tab:main}).
With 8 answer frames, recording 8 minutes plus one question costs 18.8\,s of compute, against 38.7\,s for BOLT and 67.4\,s for AKS. With 4 answer frames, the answer time halves (5.9\,s against 10.7\,s). Accuracy also rises to 59.6\%, as fewer frames gather near the peak (like Focus in \cref{tab:layout}). On an iPhone 15 Pro the same pipeline costs 18.3\,s for recording and 8.0\,s per question, with 5.6\,GB of memory at peak.
On 36 to 72 minute recordings, \method keeps an answer frame near the evidence for 44\% of questions, against 12\% for uniform sampling (\cref{sec:app_more}).

\FloatBarrier
\section{Conclusion}
\label{sec:conclusion}

We asked whether a frozen VLM can find the evidence for an egocentric question by itself.
It finds the evidence more often when it chooses among labeled frames and we read its choice from the output distribution.
\method refines this choice from coarse to fine, places a few answer frames at its quantiles, and returns the evidence time with the answer.
With Qwen3.5-4B it is more accurate than recent training-free frame selectors on QAEgo4D and MyEgo, and a single-level version runs on a Mac and an iPhone.

\paragraph{Limitations.}
Like other training-free frame selectors, \method can only find what the VLM recognizes: if the VLM does not recognize the evidence in a thumbnail, the readout cannot point to it. It also needs a VLM that takes interleaved images and returns the full probabilities of its output tokens, which most closed APIs do not.

{
    \small
    \bibliographystyle{ieeenat_fullname}
    \bibliography{main}
}

\clearpage
\setcounter{page}{1}
\onecolumn
\begin{center}
  {\Large\textbf{\thetitle}\par}
  \vspace{0.5em}{\Large Supplementary Material\par}
  \vspace{1.0em}
\end{center}
\appendix
\renewcommand{\thesection}{\Alph{section}}
\section{Proof of the quantile placement}
\label{sec:app_proof}
\paragraph{Lemma.}
Let $\mu$ be a probability measure on $[0,T]$ with distribution function $F$ and quantile function $F^{-1}(u)=\inf\{t: F(t)\ge u\}$.
Among all measures $\nu=\frac{1}{B}\sum_{b=1}^{B}\delta_{t_b}$, the 1-Wasserstein distance $W_1(\mu,\nu)$ is minimized by $t_b = F^{-1}\big(\frac{b-1/2}{B}\big)$, $b=1,\dots,B$.

\paragraph{Proof.}
In one dimension, $W_1(\mu,\nu)=\int_0^1 |F^{-1}(u)-G^{-1}(u)|\,du$, where $G^{-1}$ is the quantile function of $\nu$.
Without loss of generality $t_1\le\dots\le t_B$.
Then $G^{-1}(u)=t_b$ for $u\in\big(\frac{b-1}{B},\frac{b}{B}\big]$, so
$W_1(\mu,\nu)=\sum_{b=1}^{B}\int_{(b-1)/B}^{b/B}|F^{-1}(u)-t_b|\,du$.
The terms are independent, and the $b$-th term is $\frac{1}{B}\,\mathbb{E}|F^{-1}(U_b)-t_b|$ with $U_b$ uniform on $\big(\frac{b-1}{B},\frac{b}{B}\big]$.
An absolute deviation is minimized at a median.
Since $F^{-1}$ is non-decreasing, $F^{-1}(\mathrm{med}\,U_b)=F^{-1}\big(\frac{b-1/2}{B}\big)$ is a median of $F^{-1}(U_b)$.
The minimizers are non-decreasing in $b$, so the ordering assumption holds. $\square$

\paragraph{Remarks.}
(i) If $\mu$ is uniform on $[0,T]$, then $t_b=\frac{(b-1/2)T}{B}$, which is evenly spaced sampling.
(ii) The general result for prescribed weights is Theorem 5.5 of \cite{xu2019best}.

\section{Fewer frames}
\label{sec:app_budget}
We want to know how few answer frames \method needs, since each frame costs time on a device.
On QAEgo4D, single-level \method with 4 frames is more accurate than AKS and BOLT with 8 frames at every resolution (\cref{tab:fewer}).

\begin{table}[h]
  \centering
  \small
  \caption{\textbf{4 frames vs.\ 8 frames} (QAEgo4D, Qwen3.5-4B, \method single level). Cap: visual tokens per answer image.}
  \label{tab:fewer}
  \begin{tabular}{@{}lcccc@{}}
    \toprule
    Cap & Uniform, $B=8$ & AKS, $B=8$ & BOLT, $B=8$ & \method, $B=4$ \\
    \midrule
    88 & 48.4 & \underline{55.8} & 51.2 & \textbf{56.2} \\
    192 & 51.4 & \underline{55.8} & 54.6 & \textbf{58.0} \\
    384 & 51.2 & \underline{56.4} & 54.2 & \textbf{59.6} \\
    \bottomrule
  \end{tabular}
\end{table}

\section{Evidence between thumbnails}
\label{sec:app_between}
On QAEgo4D no thumbnail shows the evidence for 54\% of the questions, whose evidence lasts 3\,s at the median.
For these questions, the coarse-to-fine readout raises the share with at least one answer frame within 8\,s of the evidence from 45.7\% to 54.6\% (8 answer frames).

\section{Options in the query}
\label{sec:app_same}

The localization prompt of \method includes the answer options, while AKS and BOLT score frames against the question only, as in their code.
\cref{tab:app_options} removes the options from \method and adds them to the AKS and BOLT queries in the format that A.I.R.\ uses.
Without options \method stays above AKS and BOLT on every dataset, and on QAEgo4D the options do not help AKS or BOLT.

\begin{table}[h]
  \centering
  \small
  \caption{\textbf{Options in the query} (Qwen3.5-4B, 8 answer frames).}
  \label{tab:app_options}
  \begin{tabular}{@{}lccc@{}}
    \toprule
    Method & QAEgo4D & MyEgo & EgoSchema \\
    \midrule
    AKS, question & \underline{56.4} & 39.4 & 60.4 \\
    AKS, question + options & 56.0 & -- & -- \\
    BOLT, question & 54.2 & \underline{41.3} & \underline{63.6} \\
    BOLT, question + options & 52.6 & -- & -- \\
    \midrule
    \method, question & \textbf{59.0} & \textbf{42.5} & \textbf{65.8} \\
    \bottomrule
  \end{tabular}
\end{table}

\section{Temperature}
\label{sec:app_tuning}
The main tables use fixed settings for every method.
Here we vary $\tau$ for \method.
It changes accuracy by at most 0.6 points over $\tau=1$ when $\tau\le1$, and lowers it when $\tau>1$ spreads the frames (\cref{tab:tau}).

\begin{table}[h]
  \centering
  \small
  \caption{\textbf{Temperature} $\tau$ (Qwen3.5-4B, single level, 8 answer frames).}
  \label{tab:tau}
  \begin{tabular}{@{}lccccc@{}}
    \toprule
    $\tau$ & 0.25 & 0.5 & 1 & 2 & 4 \\
    \midrule
    QAEgo4D & \underline{57.2} & \textbf{57.8} & \underline{57.2} & 55.8 & 54.4 \\
    MyEgo & \textbf{42.3} & 41.7 & \underline{42.2} & 40.6 & 39.9 \\
    \bottomrule
  \end{tabular}
\end{table}

\section{Model notes and label check}
\label{sec:app_models}
\paragraph{Model notes.}
Every localization label ends with ``]'' so that no label is a prefix of another.
For every model we score label $j$ as the tokens of ``$j$]''. GLM splits this into the label token and a separate ``]'' token, so its merged \texttt{]</} token is never scored.
GLM-4.1V-9B, released as GLM-4.1V-9B-Thinking, writes prose after its default answer prefix, so its answer prefill adds its box token. The localizer prefill is unchanged.
Thinking is turned off for Qwen3.5 and MiniCPM-V-4.5.

\paragraph{Label check.}
The test shifts the labels by $N/2$ and correlates the new coarse readout with the original one, matched once by label number and once by image.
The check fires when the match by number is higher. It then reads the coarse readout once more with the labels in reverse order and combines the two by their geometric mean.
Because the check reads no answers, all results use the full test sets.
Answer letters follow the option format of each dataset, ``(A'' on QAEgo4D and ``A'' otherwise, for every method. For models that also offer the other form, we score whichever form has more total probability.

\section{Additional results}

\paragraph{Evidence coverage.}
\cref{tab:coverage} reports, for the methods of \cref{tab:main}, how often an answer frame lies within 8\,s of the evidence, and QAEgo4D accuracy by the length of the evidence window.
\method has the highest coverage on both datasets, and its lead in accuracy is largest on the briefest evidence.

\begin{table}[h]
  \centering
  \small
  \caption{\textbf{Evidence coverage} (Qwen3.5-4B, 8 answer frames). Near: share of questions with an answer frame within 8\,s of the evidence (QAEgo4D) or of the answer moment (MyEgo). Right: QAEgo4D accuracy by the length of the evidence window, annotated in whole seconds, with the number of questions.}
  \label{tab:coverage}
  \begin{tabular}{@{}lccccc@{}}
    \toprule
     & \multicolumn{2}{c}{Near} & \multicolumn{3}{c}{QAEgo4D accuracy by evidence length} \\
    \cmidrule(lr){2-3}\cmidrule(l){4-6}
    Method & QAEgo4D & MyEgo & $<$3\,s (131) & 3--10\,s (219) & $>$10\,s (150) \\
    \midrule
    Uniform & 41.6 & 49.6 & 53.4 & 47.5 & 54.7 \\
    AKS & 53.8 & 55.8 & 55.7 & 57.5 & 55.3 \\
    BOLT & 56.2 & \underline{58.0} & 54.2 & 54.8 & 53.3 \\
    A.I.R. & \underline{59.2} & 55.4 & \underline{58.0} & \textbf{62.6} & \underline{60.7} \\
    Write IDs & 55.2 & 51.7 & 56.5 & 55.3 & 56.0 \\
    FOCUS & 55.2 & 55.8 & 53.4 & 58.9 & 54.7 \\
    \midrule
    \method & \textbf{67.2} & \textbf{66.7} & \textbf{62.6} & \underline{61.6} & \textbf{62.0} \\
    \bottomrule
  \end{tabular}
\end{table}

\paragraph{Long recordings.}
A wearable records far longer than one QAEgo4D clip.
To test this, each of 100 QAEgo4D questions is asked about its own video joined with five other test videos, which gives recordings of 36 to 72 minutes (49 on average).
\method keeps $N=32$ thumbnails, so its time per question stays the same (5.95\,s against 5.82\,s of GPU time, without decoding).
It remains the most accurate, and it keeps an answer frame near the evidence almost four times as often as uniform sampling (\cref{tab:long}).

\begin{table}[h]
  \centering
  \small
  \caption{\textbf{Long recordings} (Qwen3.5-4B, 8 answer frames, 100 QAEgo4D questions). Each video is joined with five other test videos. Near: share of questions with an answer frame within 8\,s of the evidence.}
  \label{tab:long}
  \begin{tabular}{@{}lcccc@{}}
    \toprule
     & \multicolumn{2}{c}{Accuracy} & \multicolumn{2}{c}{Near} \\
    \cmidrule(lr){2-3}\cmidrule(l){4-5}
    Method & Original & Long & Original & Long \\
    \midrule
    Uniform & 51.0 & 47.0 & 43.0 & 12.0 \\
    \method, single level & \underline{52.0} & \underline{44.0} & \underline{59.0} & \underline{33.0} \\
    \method & \textbf{56.0} & \textbf{51.0} & \textbf{66.0} & \textbf{44.0} \\
    \bottomrule
  \end{tabular}
\end{table}

\paragraph{Time since the moment.}
MyEgo asks about moments that passed 15\,s to 11 minutes before the question.
\cref{tab:elapsed} splits its questions into quartiles of this gap.
\method is the most accurate or tied in every quartile, and it has a frame within 8\,s of the moment more often than any other method (Near).
\begin{table}[h]
  \centering
  \small
  \caption{\textbf{MyEgo by time since the answer moment} (Qwen3.5-4B, 8 answer frames, quartiles over 996 questions). Near: share of questions with an answer frame within 8\,s of the moment.}
  \label{tab:elapsed}
    \begin{tabular}{@{}lccccccccc@{}}
    \toprule
     & & \multicolumn{4}{c}{Accuracy} & \multicolumn{4}{c}{Near} \\
    \cmidrule(lr){3-6}\cmidrule(l){7-10}
    Gap (min) & $n$ & Uni. & AKS & BOLT & \method & Uni. & AKS & BOLT & \method \\
    \midrule
    0.25--0.37 & 254 & 35.0 & 36.6 & \underline{39.4} & \textbf{43.3} & 45.3 & \underline{64.2} & 59.4 & \textbf{73.6} \\
    0.37--0.67 & 247 & 36.8 & \underline{40.5} & \textbf{41.3} & \textbf{41.3} & 49.4 & \underline{68.0} & 65.2 & \textbf{73.3} \\
    0.67--1.60 & 246 & 34.6 & 41.1 & \underline{41.9} & \textbf{42.7} & 60.6 & 51.2 & \underline{65.9} & \textbf{67.5} \\
    1.60--10.7 & 249 & 38.2 & 39.4 & \underline{42.6} & \textbf{43.0} & \underline{43.4} & 39.8 & 41.8 & \textbf{52.2} \\
    \bottomrule
  \end{tabular}
\end{table}
\label{sec:app_more}
\paragraph{Localization input.}
\cref{tab:ablation} changes how the single-level readout sees the thumbnails.
Giving them as one video instead of labeled images costs 4.0 points, and full-resolution thumbnails cost 0.6 points and three times the localization tokens.

\begin{table}[h]
  \centering
  \small
  \caption{\textbf{Localization input} (QAEgo4D, Qwen3.5-4B, single level, 32 segments, 8 answer frames).}
  \label{tab:ablation}
  \begin{tabular}{@{}lc@{}}
    \toprule
    Variant & QAEgo4D \\
    \midrule
    Labeled thumbnails (\method) & \textbf{57.2} \\
    Full-resolution thumbnails & \underline{56.6} \\
    Thumbnails as one video & 53.2 \\
    \bottomrule
  \end{tabular}
\end{table}

\section{Timing protocol}
We time every method end to end on the same 50 QAEgo4D questions on one A100 MIG slice (3g.20gb), after one warm-up question.
Time covers decoding, preprocessing, frame scoring and selection, and answering.
For the accuracy runs, A.I.R.\ selects frames on one A100 80GB GPU, and its answers run on the same GPU as those of the other methods.

\section{On-device details}
\label{sec:app_device}
We want to know where the time goes on the device, so \cref{tab:app_stages} splits the time from question to answer (Mac, 4-bit, 6\,GB cap, the 20 questions of \cref{tab:device}, median).
Most of it is the answer pass and decoding the answer frames. Localization reads only the question against the cached thumbnails.

\begin{table}[h]
  \centering
  \small
  \caption{\textbf{Time per question on the device} (s, median, single level). Rest: question prefill on the cache, localization readout and evidence readout. Each row is a median, so the total can differ slightly from \cref{tab:device}.}
  \label{tab:app_stages}
  \begin{tabular}{@{}lccc@{}}
    \toprule
     & Mac, $B=4$ & Mac, $B=8$ & iPhone, $B=4$ \\
    \midrule
    Answer-frame decode and encode & 1.3 & 2.6 & 1.8 \\
    Answer pass & 3.7 & 7.3 & 4.7 \\
    Rest & 0.9 & 0.7 & 1.5 \\
    \midrule
    Question to answer & 5.9 & 10.6 & 8.0 \\
    Peak memory (GB) & 5.6 & 5.6 & 5.6 \\
    \bottomrule
  \end{tabular}
\end{table}

\paragraph{Readout variants.}
The localization readout scores every label sequence.
Scoring them one by one takes 627\,ms per question at 4.39\,GB peak, batching the labels takes 562\,ms at 5.38\,GB, and one pass with an fp32 cache state takes 304\,ms at 5.58\,GB.
One pass with a bf16 cache state takes 272\,ms at 4.64\,GB and picks the same top segment as scoring one by one on 50 of 50 questions.
The iPhone (iPhone 15 Pro, 8\,GB, iOS 26.6) uses this last variant, and the Mac runs in \cref{tab:device} use the fp32 state.
The iPhone caps answer frames at 192 visual tokens.

\end{document}